\documentclass[twocolumn,10pt]{asme2ej}

\usepackage{graphicx} %% for loading jpg figures
\usepackage{amsmath}
\DeclareMathOperator*{\argmax}{argmax}

\title{A DEEP REINFORCEMENT LEARNING APPROACH FOR GLOBAL ROUTING}

\author{HAIGUANG LIAO \hspace{12pt}
      {\tensfb WENTAI ZHANG} \hspace{12pt}
      {\tensfb XULIANG DONG} \hspace{12pt}\\
      {\tensfb BARNABAS POCZOS} \hspace{12pt}
      {\tensfb KENJI SHIMADA} \hspace{12pt}
      {\tensfb LEVENT BURAK KARA}\thanks{Address all correspondence to this author.}
      \affiliation{%Visual Design Engineering Laboratory\\
	Carnegie Mellon University\\
	Pittsburgh, PA 15213\\
    }
}

\begin{document}

\maketitle    

%%%%%%%%%%%%%%%%%%%%%%%%%%%%%%%%%%%%%%%%%%%%%%%%%%%%%%%%%%%%%%%%%%%%%%
\begin{abstract}
{\it Global routing has been a historically challenging problem in electronic circuit design, where the challenge is to connect a large and arbitrary number of circuit components with wires without violating the design rules for the printed circuit boards or integrated circuits. Similar routing problems also exist in the design of complex hydraulic systems, pipe systems and logistic networks.  Existing solutions typically consist of greedy algorithms and hard-coded heuristics. As such, existing approaches suffer from a lack of model flexibility and non-optimum solutions. As an alternative approach, this work presents a deep reinforcement learning method for solving the global routing problem in a simulated environment. At the heart of the proposed method is deep reinforcement learning that enables an agent to produce an optimal policy for routing based on the variety of  problems it is presented with leveraging the conjoint optimization mechanism of deep reinforcement learning. Conjoint optimization mechanism is explained and demonstrated in details; the best network structure and the parameters of the learned model are explored. Based on the fine-tuned model, routing solutions and rewards are presented and analyzed. The results indicate that the approach can outperform the benchmark method of a sequential A* method, suggesting a promising potential for deep reinforcement learning for global routing and other routing or path planning problems in general. Another major contribution of this work is the  development of a global routing problem sets generator with the ability to generate parameterized global routing problem sets with different size and constraints, enabling evaluation of different routing algorithms and the generation of  training datasets for future data-driven routing approaches.
}
\end{abstract}

%%%%%%%%%%%%%%%%%%%%%%%%%%%%%%%%%%%%%%%%%%%%%%%%%%%%%%%%%%%%%%%%%%%%%%
% \begin{nomenclature}
% \entry{A}{You may include nomenclature here.}
% \entry{$\alpha$}{There are two arguments for each entry of the nomemclature environment, the symbol and the definition.}
% \end{nomenclature}

% The primary text heading is  boldface and flushed left with the left margin.  The spacing between the  text and the heading is two line spaces.

%%%%%%%%%%%%%%%%%%%%%%%%%%%%%%%%%%%%%%%%%%%%%%%%%%%%%%%%%%%%%%%%%%%%%%
\section{Introduction}

Keeping pace with  Moore's Law, Integrated Circuits (IC) are becoming increasingly more sophisticated with the number of transistors in a unit area increasing exponentially over time \cite{kahng2018machine,schaller1997moore}. More capable automatic design systems are needed to help  engineers tasked with solving increasingly more challenging IC design problems. In the design flow of IC, global routing is a particularly critical and challenging stage in which the resources for routing (hence design constraints) in the chip design  is allocated based on the preceding step of component placement and configurations of the chip to be designed \cite{hu2001survey}. The routing problem is then addressed in two stages, the first being the global routing step (the focus of this work), followed by detailed routing where the actual path of the wires and vias connecting those electronic components are decided.

Global routing has been a historically challenging  problem in IC physical design for decades. Global routing involves a large and arbitrary number of \textbf{nets} to be routed, where each net may consist of many \textbf{pins} to be interconnected with wires.  Even the simplest version of the problem, where a single net with only two pins needs to be routed under the design constraints is  an NP-complete problem \cite{kramer1984complexity}. In real settings,  millions of components and nets may need to be integrated on a single chip, making the problem extremely challenging. Similar routing problems are encountered in other domains including routing-design of hydraulic systems \cite{hydraulic}, complex pipe system routing in ships \cite{shippipe},  urban water distribution systems \cite{christodoulou2010pipe,grayman1988modeling}, and  routing of city logistic services \cite{cityrouting,barcelo2007vehicle}. 

Owing to the difficulties of the problem and demand for tools for designing increasingly complicated IC, global routing has been one of the most active research areas in IC design \cite{hu2001survey,lee1976use,moffitt2008maizerouter,kastner2000predictable}. However, current solutions rely primarily on heuristically driven greedy methods such as net ordering followed by sequential net routing \cite{hu2001survey}, routing different areas of chip sequentially\cite{cho2007boxrouter}, and net ordering followed by force-directed routing \cite{hasan1987force}. These heuristic based methods are primarily applicable with  strong constraints on the problems to be solved. As such, there remains potential to identify better solutions that can further minimize  the objective functions of interest such as the  total wirelength and edge overflow commonly utilized in IC design.

This work presents a deep reinforcement learning (DRL) driven approach to global routing. DRL combines  reinforcement learning and deep learning, and has been successfully utilized in a variety of sequential decision making problems \cite{mnih2013playing,mnih2015human,van2016deep,kulkarni2016hierarchical,multiagent,alphagozero2018,mnih2013playing}. In our  setting, both the state space (routing states) and the action space (routing decisions) are discrete and finite. Moreover, since a future routing state only depends on the current routing state and all states features are observable, we model the problem as a \textbf{Markov Decision Process}\cite{MDP}. We choose a Deep Q-network (DQN) as our fundamental RL algorithm to address the global routing problem.

%Under our state and action design discussed in section \ref{drl}, there are only 6 distinct actions versus numerous intermediate states, which is similar to the Atari games environment Mnih et al. described in \cite{mnih2013playing}. In light of their success, 

At the heart of our approach is the observation that a properly trained \emph{Q-network} within our DQN formulation can consider the nets and the pins to be routed conjointly, as opposed to attempting to route them sequentially with little or no backtracking. A single Q-network is trained and learned on the target problem of interest. When trained, it produces a `best' action for the dynamically evolving state of the routing by taking into account all the previous successful as well as unsuccessful net routing attempts that inform a reward function.

To the best of our knowledge, the presented work is the first attempt to formulate and solve global routing as a deep reinforcement learning problem. A global routing problem sets generator is also developed to automatically generate parameterized problems with different scales and constraints, enabling comparisons between algorithms and providing  training, testing and validation data for future data-driven approaches. We describe our DQN formulation together with the parametric studies of the best parameters of the algorithm and network structure. We compare our approach against solutions obtained by the commonly used A* search, which is sequentially applied to solve a set of nets. It is noted however that our approach, similar to previous approaches, does not guarantee global optimum. 

\section{BACKGROUND and RELATED WORK}
\subsection{Global Routing}
Global routing allocates space resources on the chip to be designed  by connecting the electronic components in a circuit coarsely based on a given placement solution \cite{sechen2012vlsi}. It can be modeled as a grid graph $G(V,E)$, where each vertex $v_i$ represents a rectangular region of the chip, so-called a global routing cell (Gcell) or a global routing tile, and each edge $e_{ij}$ represents the boundary between $v_i$ and $v_j$. In a given routing problem, IC design considerations impose constraints on the edges  between neighboring global routing tiles. For this, each edge $e_{ij}$ is assigned a  capacity $c_{ij} \in \mathbf{Z^+}$  that indicates the maximum number of wire crossings that are permitted without penalty across that edge. In the final routing solution, if the number of wire crossings across $e_{ij}$ exceeds $c_{ij}$, this excess is represented as the overflow of the edge $of_{ij}$ and the real-time capacity of $e_{ij}$ becomes $-of_{ij}$.  The routing problem takes as input an arbitrary number of nets, where each net has a number of pins with known layout coordinates to be interconnected. 

\begin{figure}[t]
\begin{center}
\includegraphics[scale=0.26]{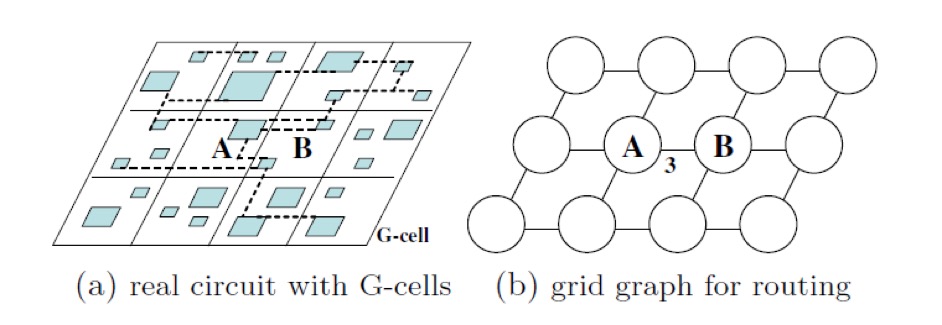}
\end{center}
% \caption{SAMPLE OF REAL CIRCUIT AND CORRESPONDING GRID GRAPH FOR GLOBAL ROUTING \cite{cho2007boxrouter}.}
\caption{Sample of a real circuit and corresponding grid graph for global routing \cite{cho2007boxrouter}.}

\label{globalroutingscheme} 
\end{figure}

Fig.~\ref{globalroutingscheme} shows an example of a real circuit and its corresponding grid graph. Global routing's goal is to find wiring paths that connect the pins inside the Gcells through $G(V,E)$ for all input nets. When connections are made through an edge whose capacity has been exhausted, the excess crossings are counted  as the \emph{overflow} (OF) for that edge. The goal of global routing is to minimize:  (1) The \emph{congestion} in the final design, which is the  total overflow accrued over all the edges, and (2) The total \emph{wirelength} (WL), which is the sum of tile-to-tile connections emanating from all the nets used in the design. If two routing solutions achieve the same capacity constraint, the one with smaller wirelength will be a better solution.  In most cases, a grid graph to be routed consists of more than one layer, and the connection from one cell to the neighboring cell in the layer above or below is called a via. 

%In a given routing problem, IC design considerations impose constraints on the edges  between neighboring global routing tiles.  For each edge, this information is encoded as the \emph{edge capacity} that specifies the maximum number of wires that can pass through the edge.  All nets must be routed within the global routing graph, ideally with no graph edges exceeding its capacity. 

%xxx: update this paragraph based on what I wrote above. Some of what is below is now redundant now. Try to minimize the use of adjectives and adverbs, you seem to use words such as 'some, essentially, literally etc.' try not to use these, it does not look very professional. 

%In order to run, test and compare different global routing algorithms, benchmarks, which are toy grid graphs for routing, is developed. 

In global routing, the nets are usually routed sequentially \cite{hu2001survey}. In routing each single net,  there are often more than two pins to be connected, in which case  methods such as the minimum spanning tree (MST) or rectilinear Steiner tree (RST) is used for decomposing a multiple-pin net problem into a set of two-pin connection problems \cite{hu2001survey,de1998rsr,ho1990new,wong1990global}.  Once a net is decomposed into a set of two-pin connection problems, search algorithms such as A*  is used to find a solution to the two-pin problems. Once the set of two-pin problems of a net are solved, these solutions are merged into a solution for a single net. 
 
Based on the above primary solution strategies, more advanced heuristic-based techniques including rip-up and reroute \cite{cho2007boxrouter2}, force-directed routing, and region-wise routing \cite{hu2001survey} have been developed to increase the performance of global routing techniques. Besides, some machine learning based techniques has been applied to solve global routing such as prediction of routing congestion models \cite{qi2014accurate} or routability prediction \cite{zhou2015accurate} with supervised models. However, these machine learning based approach only solves part of the problem instead of providing a closed loop global routing solution. Also, to our knowledge, no published research of global routing has investigated deep reinforcement learning as a possible new strategy, which might well fit for solving Markov Decision Process like global routing. 

\subsection{Deep Reinforcement Learning}

%xxx: open this section with a proper introduction. just having an equation is too abrupt. Also, here and all the other euations that are introduced, we need to explain all the variables usedin the equations. This is expecvially critcal for equations 1,2,3, 4 but applies to everything that comes after those too. See my siggraph papers for exampels on how they shoyld be introduced. 

At the heart of reinforcement learning is the discovery of the optimal action-value function $Q^*(s,a)$ by maximizing the expected return starting from state $s$, taking action $a$, and then following policy $\pi$ from there on that gives rise to the subsequent action-state configurations \cite{sutton1998introduction}:

\begin{equation}
    Q^*(s,a) = max_{\pi} E[R_t|s_t=s,a_t=a,\pi] 
\label{Qfunction}
\end{equation}

The above optimal action-value function  obeys the \textit{Bellman equation} (Eqn.~\ref{Bellman}), wherein many reinforcement learning algorithms estimate the action-value function in an iterative manner.  Obtaining updated value of action-value function following $Q_{i+1}(s,a)=E[r+\gamma max_{a'}Q^*(s',a')|s,a]$, with expectation of current reward $r$ and maximized future reward multiplied with a discount factor $\gamma$, given a current state $s$ and an action $a$ taken from that state. 
\begin{equation}
    Q^*(s,a) = E_{s'\sim\xi}[r+\gamma max_{a'}Q^*(s',a')|s,a]
\label{Bellman}
\end{equation}

In actual implementation, a function approximator $Q(s,a;\theta)$ is used to estimate the optimal action-value function. In DRL, a deep neural network such as a convolutional neural network (CNN) \cite{mnih2013playing} or fully connected networks are used as the function approximator. The neural network used in the algorithm is often referred to as the Q-network. The Q-network is trained with a batch of sequences and actions by minimizing loss functions $L_i(\theta_i)$, which can be obtained by Eqn.(\ref{lossfunction}). Here, $y_i=E_{s'\sim\epsilon}[r+\gamma max_{a'}Q(s',a';\theta_{i-1})|s,a]$ is the target for iteration $i$ and $\rho(s,a)$ is a probability distribution over the states and actions. 

By differentiating the loss function with respect to its weights $\theta_i$, the gradient Eqn.(\ref{gradient}) is obtained. Based on the gradient, the network is trained with stochastic gradient descent. Once trained, actions can be obtained from the network following an $\epsilon$ greedy strategy \cite{mnih2013playing}.

\begin{equation}
    L_i(\theta_i) = E_{s,a\sim\rho(.)}[(y_i-Q(s,a;\theta_i))^2]
\label{lossfunction}
\end{equation}

\begin{equation}
    \nabla_{\theta_i}L_i(\theta_i) = E[(r+\gamma
    max_{a'}Q(s',a';\theta_{i-1})-Q(s,a;\theta_i)) dQ]
\label{gradient}
\end{equation}

\begin{equation}
    d Q = \nabla_{\theta_i}Q(s,a;\theta_i)
\label{gradient2}
\end{equation}

%i commented out the paragraph below. We should minimize/avoid any comparions or analogies to atari or AGZ. I too find that to be too simplistic of a comaprison that is dangerous and we had a comment about this in the idetc paper.

%In recent years, there have been a multitude of examples that demonstrate DRL can outperform humans in challenging problems  such as the game of AlphaGo Zero \cite{alphagozero2018} and Atari game player \cite{mnih2013playing}.  These problems tend to be very high dimensional and involve sequential decision making. Inspired by such works, in this work we make a first proof-of-concept attempt to apply DRL to solve the global routing problem. 

%%%%%%%%%%%%%%%%%%%%%%%%%%%%%%%%%%%%%%%%%%%%%%%%%%%%%%%%%%%%%%%%%%%%%%
\section{METHOD}
\label{drl}
\begin{figure}[t]
\begin{center}
\includegraphics[scale=0.55]{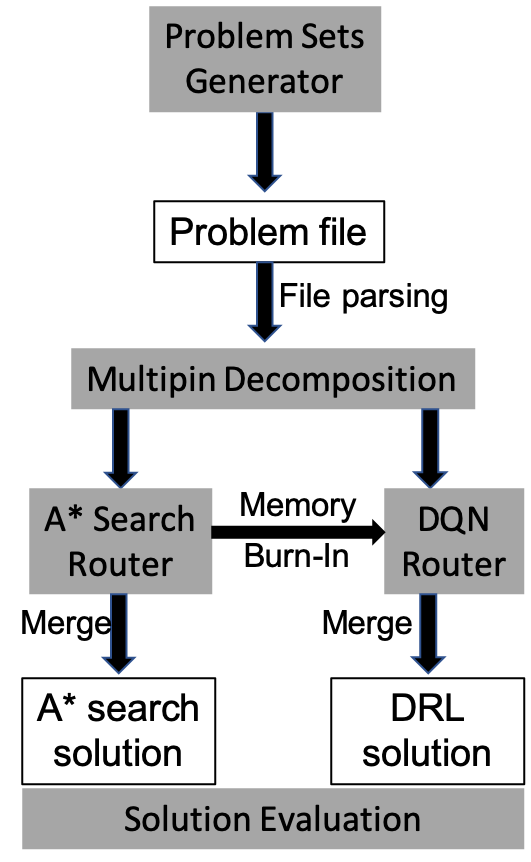}
\end{center}
\caption{Pipeline for solving global routing with DQN.}
\label{pipeline} 
\end{figure}

Figure \ref{pipeline} illustrates our DQN-based approach to global routing. First, the problem sets generator generates problems with a specified size, complexity and constraints. All the problems generated are stored in separate text files. Following the convention in the global routing problems, the input text file contains all the information necessary to describe the  problem, including the size of the grid, the edge capacity of the edges in the grid, the nets to be routed, and for each net the spatial $x,y,z$ coordinates of the pins that comprise the net. The input file is read in and parsed. Then, each  net is further decomposed into a set of two-pin problems with the minimum spanning tree algorithm. A*  router and DQN router are then used to solve the large set of two-pin problems.  A*  is executed first in order to provide burn-in memory for the DQN solver. After all two-pin problems emanating from all the nets are solved, the solutions belonging to the individual nets are merged. After a final solution is obtained for the entire set of nets, it is evaluated in light of the total congestion (sum of edge overflows, OF) and the total wirelength (WL) to assess the quality of that particular solution.  All steps except solution evaluation are programmed in Python 3.6, and the evaluation is done using a separate module written in Perl. The details of the various steps are provided next.

\subsection{Environment Setup}
Given a target chip routing problem, the text description input files consist of the chip information and all the nets to be routed. Chip information consists of grid size, edge capacities and the reduced capacities of some of the edges owing to the pre-routed nets or other obstacles present on the chip. The nets information consists of  a list of pin locations in each net. For a standard large scale target problem, the number of pins belonging to a single net is not equal and may, in fact, vary from as few as two to more than a thousand. After the input  file is read in and parsed, a simulated global routing environment that incorporates all the chip  and nets information is created. The simulated environment is designed to have the following functions:

\begin{enumerate}
    \item For each net, decompose multi-pin problems into a set of two pin problems. Feed routing solver with sets of two-pin routing problems, store and merge two-pin solutions to final global routing solution.
    
    \item Provide reward feedback and observed sequence to DQN algorithms.
    
    \item Iteratively update edge capacities.
\end{enumerate}

%xxx: I don't quite understand the use of the word benchmark and I am removig those in this paper. Benchmark implies that it alredy has a solution and we would be comparing our to it. Our sitution is not like that. We still solve the problems using A*. I changed the text so that I don;t use benchmakrs anymore, With that,  in the above, I don;t underatnd what Update capcity of becnhmarks means. Did you mean capacity of the edges?

%[yyyexplain]: Yes, by saying capacity benchmark, I mean the capacity of edges. I have also replaced existing "benchmark" with words "sample problem" or "problem"

\begin{figure}[t]
\begin{center}
\includegraphics[scale=0.4]{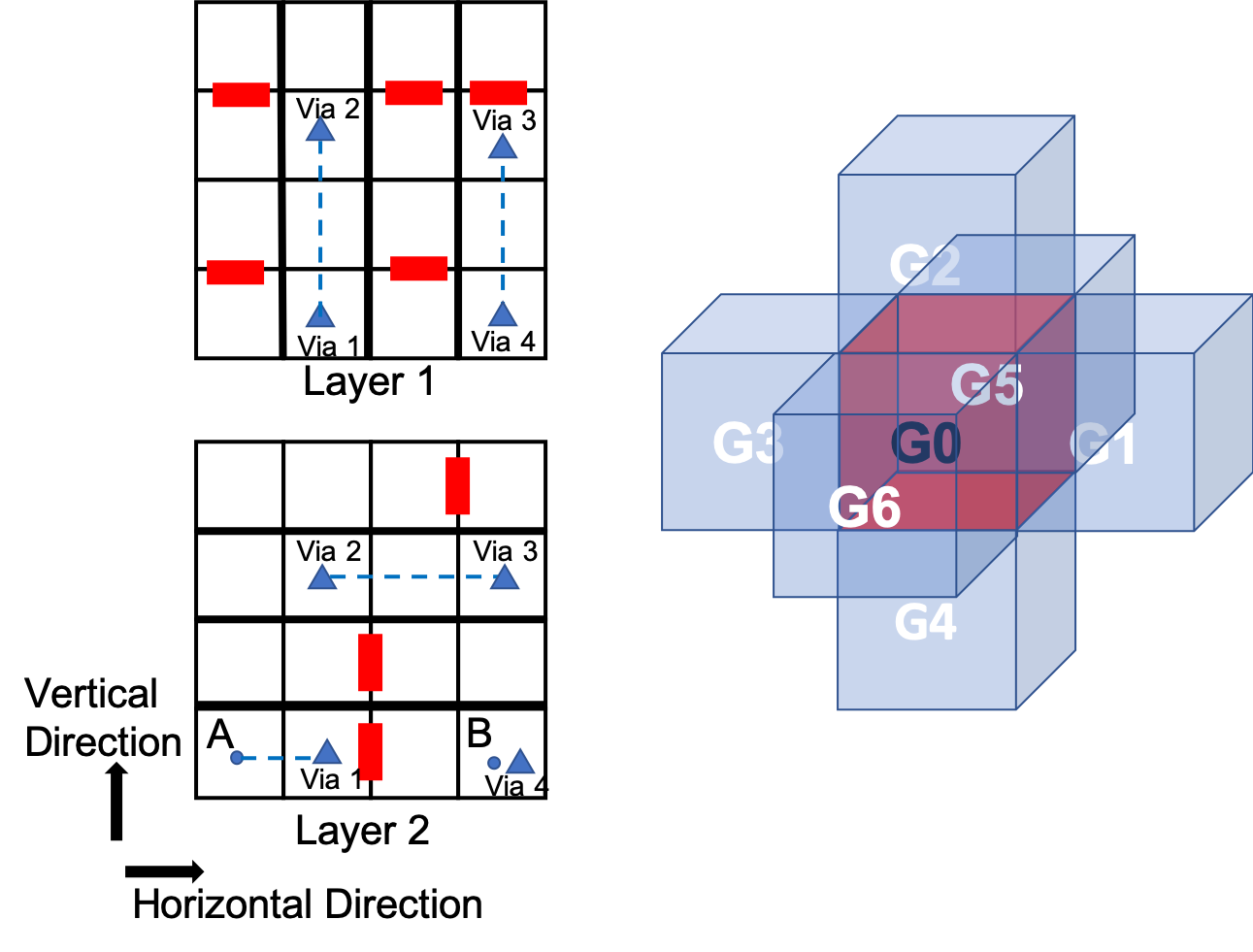}
\end{center}
\caption{Example configurations of the simulated global routing environment.}
\label{gridconfigure} 
\end{figure}

In general, the simulated global routing environment can be compared to a sequential version of a maze game, with  edge capacities changing dynamically according to the thus-far routed path. Fig.~\ref{gridconfigure} shows an example of a simulated two-layer ($4\times 4\times 2$) problem environment with an illustrative routing solution for a two-pin problem. In Fig.~\ref{gridconfigure}, each layer consists of 16 Gcells ($4\times 4$), with Layer 2 stacked right above Layer 1. Bold edges have zero capacity, therefore the route can only be north-south (referred to as the vertical direction) in Layer 1 and east-west (referred to as the horizontal direction) in Layer 2 without generating OF. Red rectangles represent blocked edges owing to pre-routed wires or the existence of certain components. The two-pin problem is to generate a route from pin A to pin B, through different Gcells with the least WL and least OF possible. At each step, starting from a Gcell, there are  6 possible  moves, as shown in Fig.~\ref{gridconfigure} (right), which are going east (G1), west (G3), north (G5), south (G6), up (G2) and down (G4). In the simulated environment, actions are subject to boundary conditions and capacity constraints, therefore actual possible actions will vary and can be less than 6, depending on the environment settings of the problem. An illustrative solution to the problem is shown in Fig.~\ref{gridconfigure}. Once a  routing step is taken, the capacity of the crossed edge changes accordingly. 

In this work, $8\times 8\times 2$ problems consisting of different net numbers, pin numbers and capacity are used to explore different settings and parameters. Results on larger problems are also shown to demonstrate scalability. 

% including evalution part

\subsection{Global Routing Problem Sets Generator}
In order to facilitate comparison of different algorithms of different scale and constraints, a global routing problem sets generator is developed with the ability to automatically generate a large problem database with user-specified \emph{problem numbers, grid size, number of nets, number of pins each net, capacity settings}. The generator imitates the features of industrial global routing problems by breaking down the capacity settings into two parts: normal capacity and reduced capacity. Normal capacity determines the capacity of edges on horizontal direction ($x$ direction) for the first layer and vertical direction ($y$ direction) on the second layer. Following the properties of global routing problem, via capacity is set to be large enough in the simulated environment so that overflow will not occur in the via direction ($z$ direction) and is therefore not specified in the problem file. Reduced capacity is described by setting a set of specific edges in the problem to prescribed capacity values. In our problem sets generator, reduced capacity can be set based on edge utilization conditions, which are statistical results showing the traffic conditions of all edges based on solutions given by A*. For instance, users can set a proportion of the most congested edges to generate a problem with very stringent constraints. 

\begin{figure*}[thpb]
\centering
\includegraphics[width=\textwidth]{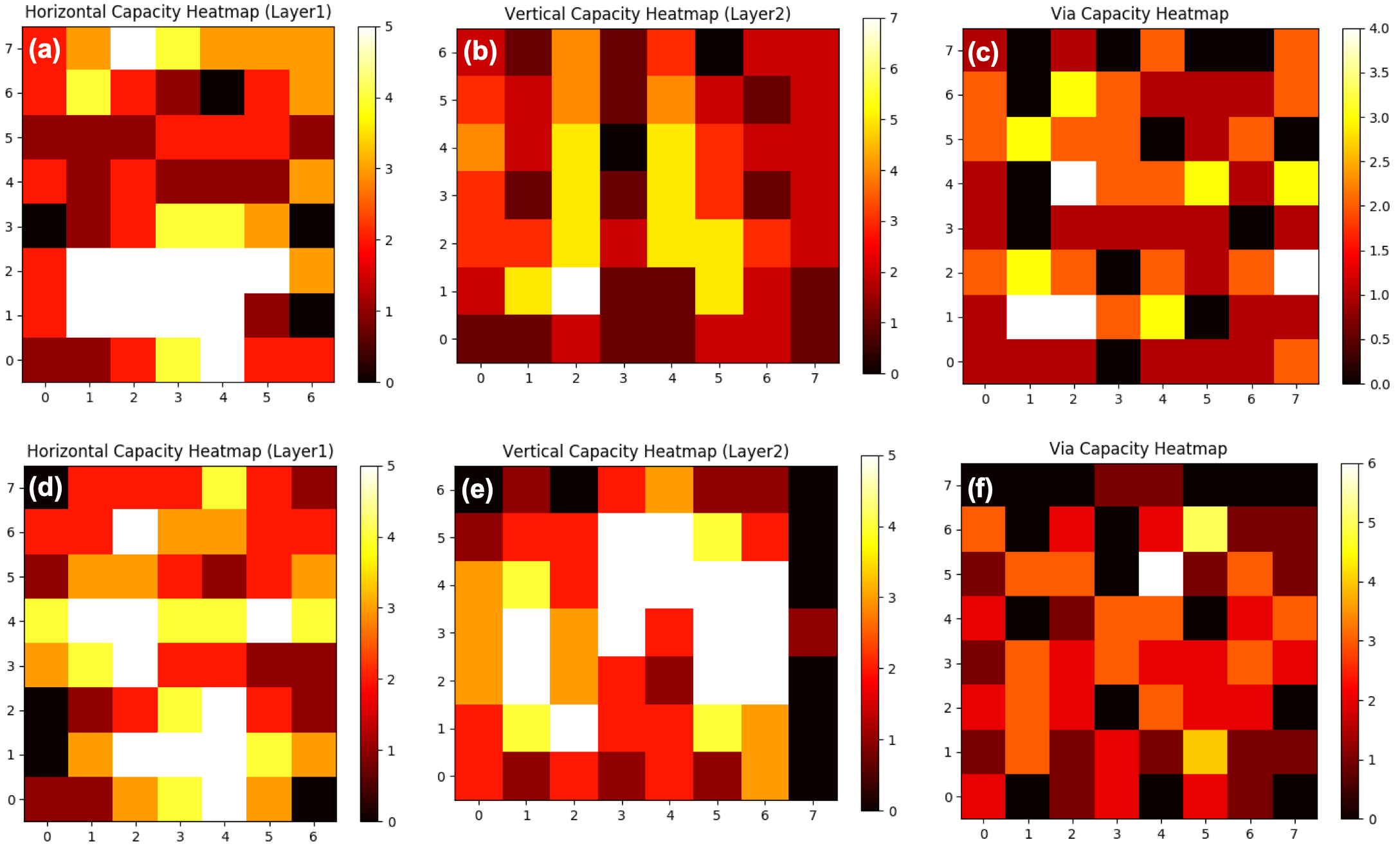}
\caption{Heat maps and edge utilization histogram of two global routing problems generated with global routing problem set generator.  $8\times 8\times 2$ benchmark with normal capacity setting: (a) via capacity heat map, (b) vertical capacity heat map, (c) horizontal capacity heat map, (d) edge utilization histogram;  $8\times 8\times 2$ benchmark with reduced capacity setting: (e) via capacity heat map, (f) vertical capacity heat map, (g) horizontal capacity heat map,(h) edge utilization histogram. In these plots, darker cells correspond to less capacity utilization while lighter colored cells correspond to higher capacity utilization. (number of nets: 50, max number of pins each net: 2, normal capacity: 3)}
\label{heat8by8} 
\end{figure*}

Fig.~\ref{heat8by8} provides an example of edge utilization conditions of two generated problems with parameters specified (gridsize: $8\times 8\times 2$, number of nets: 50, max number of pins each net: 2, normal capacity: 3). The first problem (first row) does not have reduced capacity, while the second one (second row) got three of its most congested edge capacities reduced. The heat maps demonstrate the edge utilization condition of these two problems based on A* solutions, and the difference in the traffic condition of the problems can be observed. This is an example to demonstrate the problem sets generator's ability to generate problems with significant variability in not only general problem settings such as size and normal capacity but also detailed settings such as the amount and distribution of pin numbers and locations. More examples of heat maps based on different generated problem sets are provided in Appendix A. The source code for problem sets generator will be made publicly available.

\subsection{Problem breakdown and A* Search Solver}
The solution obtained by A* serves as a benchmark against which the DRL solution can later be compared in terms of total WL and OF. In order to reduce the dimensions of the problem, a sequential A* approach is adopted, as shown in Fig.~\ref{pipeline}. All the nets in a single problem are dealt with sequentially, during which nets with multiple pins are decomposed with the minimum spanning tree method mentioned above. 

%[yyyexplain]:(above correction is to address question below: I made a brief reminder of pin decomposition problem to make the method introdtucion coherent)
%xxxx: isn't the following too much repetition? We already said these things: For a single net, if it consists of more than two pins, a standard minimum spanning tree is used to decompose the multi-pin problems into a set of two-pin problems. In multi-pin decomposition, capacity of the benchmark is not taken into account.

Before DQN router starts to work, an A* search router is used to solve those two-pin problems.  A* finds a path between the two pins (one of them taken as start $S$, the other taken as goal $G$), with the cost function defined as:

\begin{equation}
    f(n) = g(n) + h(n)
\label{A*cost}
\end{equation}

\noindent where $n$ represents the time step, $g(n)$ represents the total path length  from $S$ to the current position, and  $h(n)$ represents the future heuristic cost from the current position to $G$. 

For each step, if there is no OF, the transition cost (incremental path length) is 1. In case of an OF, the cost becomes 1000. For $h(n)$, we use the Manhattan distance from the current position to $G$,  which makes the heuristic admissible and thus the result optimal for that particular two-pin problem. However, it must be noted that the optimality of A* for the two-pin problems does not imply that the eventual final A*  solution will be globally optimal, as the nets (and hence each net's resulting two-pin problems) are solved sequentially rather than conjointly. This issue is discussed in detail in the Results and Discussion section.

As shown in Fig.~\ref{pipeline}, in addition to serving as a benchmark, A* search also provides   burn-in memory for DRL to allow DRL to converge faster. However, using A* as burn-in memory is optional, as early random walks could also be used in place of A* for burn-in memory. Nonetheless, as will be shown later, our results indicate that using A* as burn-in for DRL allows DRL to converge much faster. An experimental approach is applied to determine the best for burn-in memory. 
%wwww: check

\subsection{Deep Q-networks Router Implementation}
\begin{figure}[t]
\begin{center}
\includegraphics[scale=0.5]{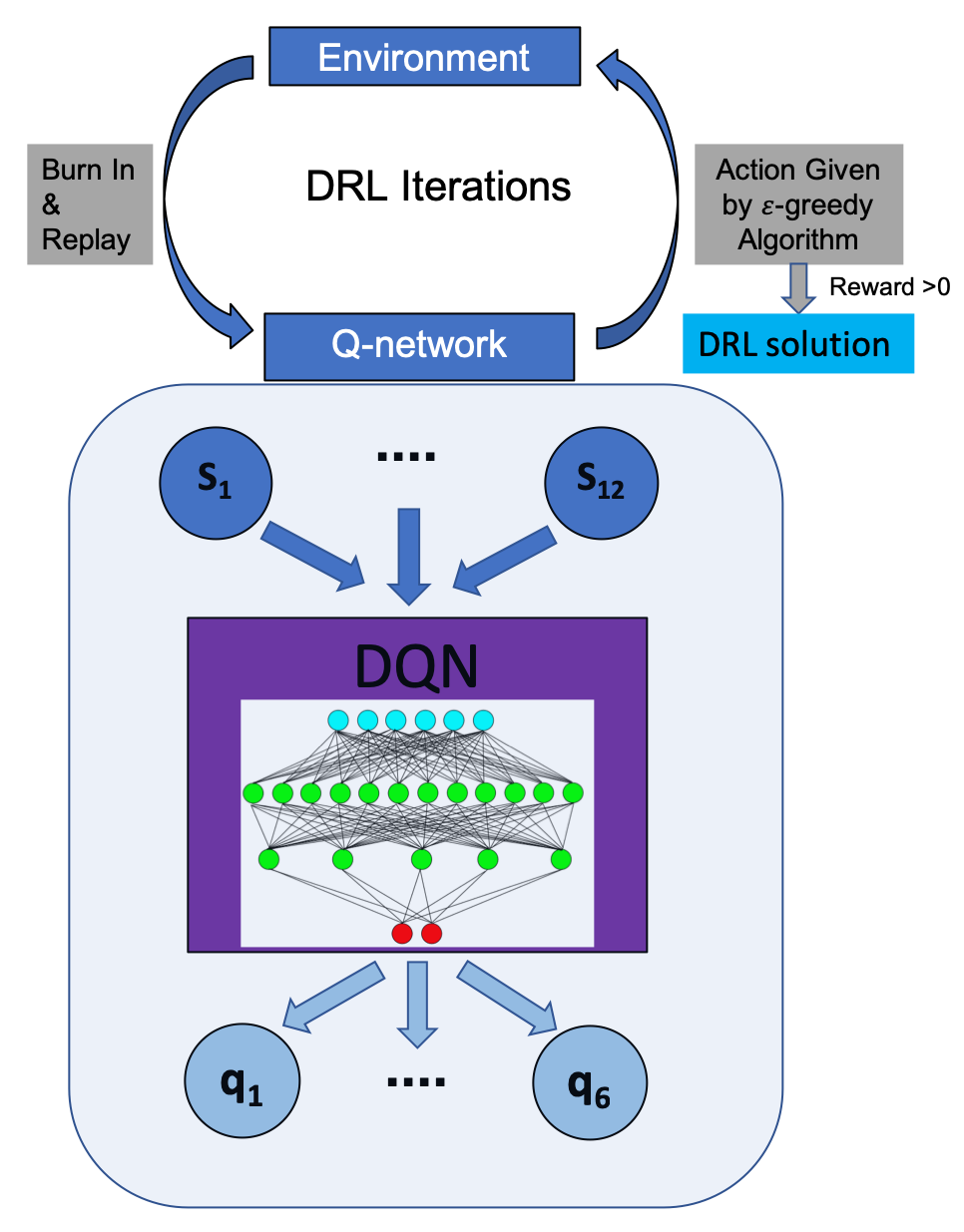}
\end{center}
\caption{Workflow of DQN-based router.}
\label{pipelineDRL} 
\end{figure}

Using the same problem breakdown, a deep Q-Network is utilized to  solve all the two-pin problems in an iterative learning process. The framework of DQN  is illustrated in Fig.~\ref{pipelineDRL}. In this framework, the Q-Network functions as an agent interacting with the  environment. Specifically, for each two-pin problem, the environment provides the network with state information. Then the agent evaluates the Q values of all the potential next states $(q_1, q_2 ... q_6)$. Finally, based on an $\epsilon$-greedy algorithm, an action is chosen and executed, altering the  environment. The agent will have thus taken one ``step'' in the environment. A reward is calculated according to the new state and  the edge capacity information is updated. In parallel, a replay buffer  records each transition along the training process. These transitions are  used for backpropagation when iteratively updating the weights in the Q-network. Further details of the  DQN routing algorithm can be found in Appendix B. Critical elements of our approach are presented below.

% capacity definition?

\textbf{State design:} The state is defined as a 12-dimensional vector. The first three elements are the $x,y,z$ coordinates of the current agent location in the environment. The fourth through the sixth elements encode the distance in the $x$, $y$ and $z$ directions from the current agent location to the target pin location. The remaining six dimensions encode the capacity information of all the edges  the agent is able to cross. This encoding scheme can be regarded as an admixture of the current state, the navigation and the local capacity information. 

\textbf{Action space:} The actions are represented with an integer  from 0 to 5 corresponding to direction of move from the current state. 

\textbf{Reward design:} The reward is  defined as a function of the chosen action and the next state $R(a, s')$. In our case:
\begin{equation}
 R(a, s')=\begin{cases}
+100&\text{if $s'$ is the target pin},\\
-1&
\text{otherwise}.
\end{cases}   
\end{equation}

\noindent This design forces the agent to learn a path as short as possible since any unfruitful action will cause a decrement in the cumulative reward. Additionally, we limit the maximum number of steps the agent can take when solving each two-pin problem to be less than a maximum threshold $T^{max}$, depending on the size of the problems to be solved. Following this reward design, the cumulative reward is always between $(100-T^{max})$ and 100 when the two-pin problem is solved, and $-T^{max}$ otherwise. This scheme is a useful indicator to distinguish if the overall routing problem was successfully solved, or no feasible solution was found. Feasible $T^{max}$ needs to be tuned according to the problem size.

%xxx: what did we use for Tmax for the 8X8 cases? [yyyexplain]: (we also used 20 for 8by8, I have added this above)

\textbf{Replay buffer and burn-in:} The replay buffer is an archive of  past transitions the agent experiences and is updated during  training. A burn-in pre-process is introduced to fill in the replay buffer before  training  begins, which provides a basic knowledge of the environment to the network. In our case, the burn-in transitions are acquired during  A* search. In each training iteration, a batch of transition records are randomly sampled from this replay buffer and used to update the network weights through backpropagation. 

As the Q-network is being trained, the two-pin problems that are solved need not be completed. That is, the network uses unsuccessful attempts (negative reward) at connecting the pin pairs in addition to the successful attempts (high, positive reward). 

%[yyy]The experience replay with random sampling is able to alleviate the bias resulting from the strong correlation among the time-sequential transitions.

\textbf{$\epsilon$-greedy algorithm:} In our case, the $\epsilon$-greedy algorithm is utilized in the following form:
\begin{equation}
A(s)=\begin{cases}
\text{a random action}&
\text{with $\epsilon$ chance}\\
\argmax_{a\in A} Q(s,a) & \text{with $1-\epsilon$ chance}.
\label{epsilongreedyeqn}
\end{cases}   
\end{equation}

\textbf{Network Architecture: }The Q-network consists of three fully-connected layers with 32, 64 and 32 hidden units in each layer. Each layer is followed by a ReLU activation layer \cite{nair2010rectified}. The input size is 12, which is the same as the number of elements in the designed state vector. The output size is 6, aligned with the number of  possible next states.

\textbf{Training Episodes:} Given the nets and the associated extracted pin-pairs of each net, the Q-network gets updated iteratively over many cycles on the same target problem. A single passage of the entire set of pin-pairs, collected over all the nets, through the network constitutes an \emph{episode}. The network is trained by processing the same pin-pairs over many episodes until convergence is reached. Once converged, the output of the Q-network allows the determination of the best action (as deemed by the network) for the given state of the problem. 

\textbf{Training Parameters: }These parameters are listed in Tab.~\ref{train_para}, which only includes the optimized set of parameters. The model and parameter tuning process are shown and discussed  next.

\textbf{Training Specifications: }The implementation of the proposed algorithm is completed with Python and Tensorflow. The training process of an $8\times 8 \times 2$ problem usually takes an hour on a workstation with an Intel Core i7-6850 CPU and an NVIDIA GeForce GTX 1080 Ti GPU.

\begin{table}[t]
\caption{DQN model optimized parameters.}
\begin{center}
\label{train_para}
\begin{tabular}{c l l l}
\hline
Parameter & Value  & Parameter & Value\\
\hline
Learning rate & 1e-4 & Batch size & 32\\
Buffer size & 50000 & Burn-in size & 10000 \\
$\gamma$ (Discount factor) & 1.0 & Burn-in memory & A*-based \\
$\epsilon$ & 0.05 & Max episodes & 5000 \\

\hline
\end{tabular}
\end{center}
\end{table}

% XXX there are 12 input nodes and 6 output nodes. Change the number in the figure. You can use ... in between.
%[yyyexplain] : (above suggestion done)

\section{EXPERIMENTS}
Firstly, to make DQN models work, general parameter sets are selected for the batch size (32, 64, 128), learning rate (1e-3, 1e-4. 1e-5), different network structures and $\epsilon$ values (0.02, 0.05, 0.10).  

In order to help better understand the conjoint optimization mechanism of DQN routing and facilitate tuning of the model, DQN models with different parameters are trained and then  applied to solve a number of parameterized routing problems generated with Problem Sets Generator. Table.~\ref{exp_num} lists the parameters of generated problems and DQN model parameters in the experiments. In the table, \emph{net numbers, capacity (global), max pin numbers (each net)} belong to problem parameters; while \emph{episodes, max step, $\gamma$, $\epsilon$, burn-in} belong to DQN model parameters. In order to demonstrate statistically significant results, each case is run on 40 problems, with the location of the pins and the number of pins randomly selected. In order to simulate reduced capacity specifications, the capacity for three of the most congested edges evaluated based on A* solutions are reduced. 

\begin{table*}[thpb]
\centering
\caption{Problem and model parameters of DRL routing experiment.}
\label{exp_num}
\begin{tabular*}{\textwidth}{l c c c c c c c c c}
\hline
No. & Net Number & Capacity & Max Pin Number & Episodes & Max Step & $\gamma$ & $\epsilon$ & Burn In & DRL Win (\%) \\
\hline
1 & 50 & 5 & 2 & 5000 & 50 & 0.98 & 0.05 & A* & 77.5\\
2 & 50 & 5 & 2 & 5000 & 50 & 0.95 & 0.05 & A* & 90\\
3 & 50 & 5 & 2 & 5000 & 50 & 0.9 & 0.05 & A* & 97.5\\
4 & 50 & 5 & 2 & 5000 & 50 & 0.8 & 0.05 & A* & 95\\

5 & 50 & 5 & 2 & 5000 & 50 & 1 & 0.05 & A* & 80\\
6 & 50 & 5 & 2 & 5000 & 50 & 1 & 0.05 & Random & 57.5\\
7 & 50 & 5 & 2 & 5000 & 50 & 1 & 0.05-0.01 & A* & 77.5\\
8 & 30 & 5 & 2 & 5000 & 50 & 1 & 0.05 & A* & 22.5\\

\hline
\end{tabular*}
\end{table*}

%%%%%%%%%%%%%%%%%%%%%%%%%%%%%%%%%%%%%%%%%%%%%%%%%%%%%%%%%%%%%%%%%%%%%%
\section{RESULTS AND DISCUSSION}
\subsection{Conjoint Optimization of DQN routing}
The main advantage of the Q-network-based DRL  for routing is the model's ability to  consider subtasks (nets and pins to be routed) conjointly, which in this paper is referred to as \textbf{conjoint optimization} of the DQN model. In exploring the advantage of such a mechanism, a study of routing problem types is conducted beforehand. 

Based on a series of experiments on the generated problems with different parameters, one key difference between the problems is based on how the edges are utilized in  A* routing. In the first category, no edge that originally has a positive capacity is fully utilized based on  A*  and this category of problems is referred to as a \textbf{Type I problem} or \textbf{no-edge-depletion problem}. In the second category, at least some edges that originally have a positive capacity are fully utilized based on  A*, and this category is referred to as a \textbf{Type II problem} or \textbf{partial-edge-depletion problem}. 

\begin{figure*}[thpb]
\centering
\includegraphics[width=\textwidth]{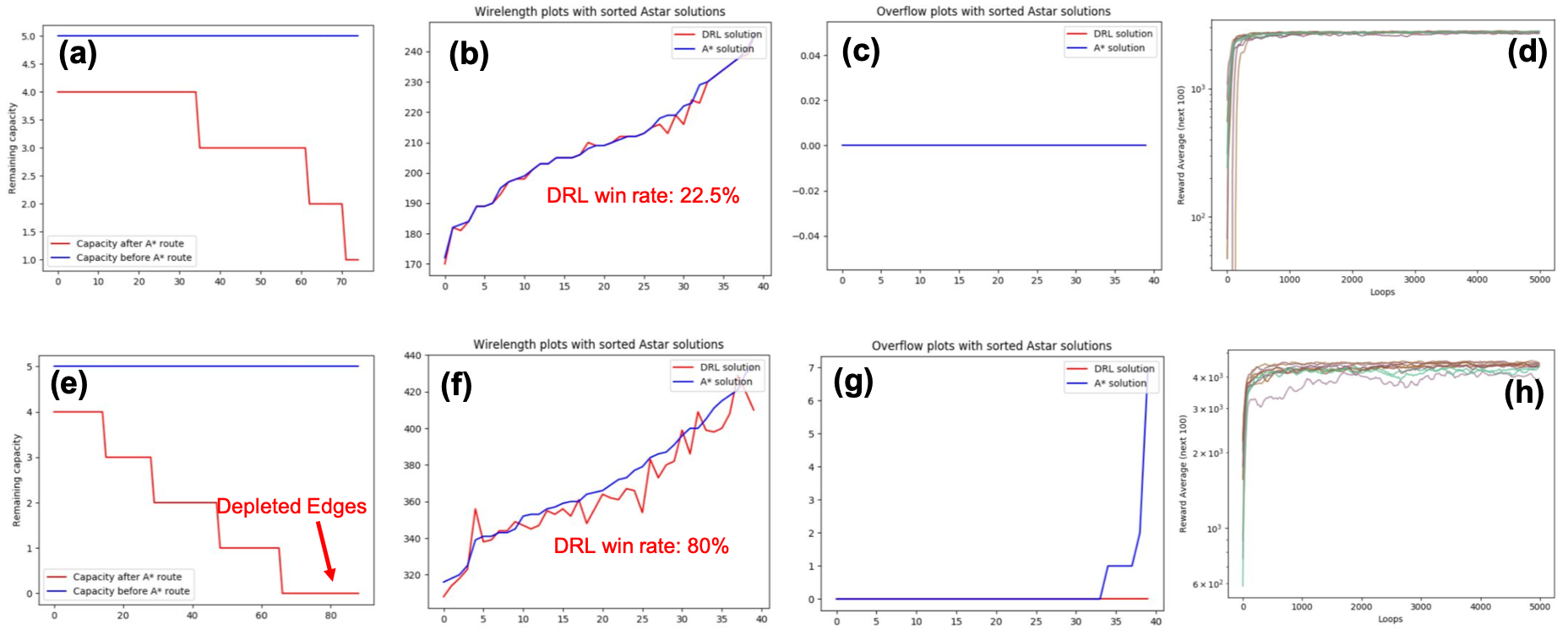}
\caption{Analysis on DQN model performance on two types of problems. Type I problem: no-edge-depletion type: (a) edge utilization plot; comparison of (b) WL, (c) OF based on DQN routing solution and A* solution, (d) log-plot reward curves of 10 cases. Type II problem: partial-edge-depletion type: (e) edge utilization plot; comparison of (f) WL, (g) OF based on DQN routing solution and A* routing solution, (h) log-plot reward curves of 10 cases.}
\label{dqnfavor} 
\end{figure*}

Type I and Type II problems conditions are simulated by experiment parameter No.5 (Type II) and No.8 (Type I) respectively. The only different parameter for the two sets of problems is the net number: in experiments No.5 there are 50 nets in each problem and in experiments No.8 there are  30 nets in each problem. Under such parameter settings, Type II problems appears more often in experiments No.5 then those in No.8. Fig.~\ref{dqnfavor} shows the analysis of the DQN models performance on two types of problems as well as the plots showing edge utilization conditions of the two types of problems based on A* solutions. Fig.~\ref{dqnfavor}a and Fig.~\ref{dqnfavor}e show the edge utilization plots of edges before (blue) and after (red) A*  for a Type I (Fig.~\ref{dqnfavor}a) and a Type II (Fig.~\ref{dqnfavor}e) problem. For Type I problems, with fewer nets to be routed, no edges are fully utilized after A*; while for Type II problems, more nets need to be routed within the same space. As such, some edges are fully utilized after A*, as indicated by the red arrow in Fig.~\ref{dqnfavor}e. 

Fig.~\ref{dqnfavor}b and Fig.~\ref{dqnfavor}c show the comparison of WL and OF based on DQN routing solutions (red) and A* (blue) on 40 cases of No.8 experiments (Type I dominant); while Fig.~\ref{dqnfavor}f and Fig.~\ref{dqnfavor}g show the comparison of WL and OF based on DQN solutions (red) and A* (blue) on 40 cases of No.5 experiments (Type II dominant). By comparing the performance of DQN routing solutions on the two types of problems, it can be seen that the DQN router outperforms A*  in most cases (80\% in terms of WL) in Type II dominant cases, while in Type I dominant cases DQN router only outperforms A* in less than one third of the cases (22.5\% in terms of WL). 

\begin{figure}[t]
\centering
\includegraphics[scale=0.7]{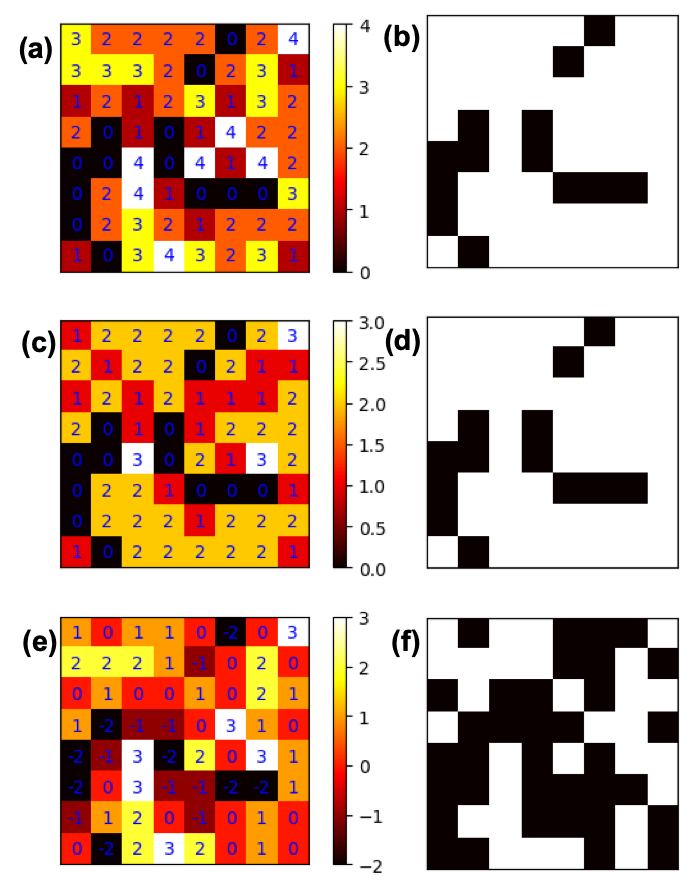}
\caption{Example capacity and their binarized plot. (a),(b): original capacity; (c),(d): no-edge-depletion capacity; (e),(f): partial-edge-depletion capacity.}
\label{capacityExplain} 
\end{figure}

The main reason for the difference in DQN router performance lies in the change of edge capacity when routing nets sequentially. To better illustrate this, some example capacity configurations are provided in Fig.~\ref{capacityExplain}, which shows illustrative capacity configurations of an original state (Fig.~\ref{capacityExplain}a and Fig.~\ref{capacityExplain}b), no-edge-depletion state (Fig.~\ref{capacityExplain}c and Fig.~\ref{capacityExplain}d) and partial-edge-depletion state (Fig.~\ref{capacityExplain}e and Fig.~\ref{capacityExplain}f). One can think of this as the capacity heat map in one of the three directions (horizontal direction, vertical direction and the via direction) of a global routing problem. The no-edge-depletion state and partial-edge-depletion state are generated by occupying a proportion of the capacity of the original state, thus simulating the capacity condition during routing after a subset of the nets and a subset of the pins have been routed. Binarized plots are capacity plots thresholded at zero: grids with capacity equal to or below zero (corresponding to an overflow condition) are marked in black, while grids with positive capacities are marked in white. 

For a Type I (no-edge-depletion) problem, since no edge is totally utilized after routing all nets, when routing the nets and pin pairs sequentially, the capacity configuration will be similar to the case in Fig.~\ref{capacityExplain}c throughout the routing procedure. In this way, even though the capacity is changing after some of the nets and pins have been routed, the binarized capacity heat map would not change throughout the process. For  A*, it only matters if a certain edge has  a positive capacity (not-blocked) or zero capacity (blocked). In other words, A*  finds solutions of a two-pin problem based on the binarized capacity heat map. With admissible heuristics, which is the case in this research, A* router is guaranteed to yield the optimum solution for each two-pin problem. Since the binarized capacity heat map would not change throughout the process in solving a Type I problem, A* solutions are guaranteed to provide the globally optimum solution for the entire routing problem given a pin decomposition algorithm for the following reasons: firstly, pre-routed nets will have no influence on the nets routed later, since the binarized capacity conditions are not changing; secondly, a routing problem solution is essentially a combination of solutions for each two-pin problems. Thus, in solving a Type I problem, DQN router will not necessarily exhibit any advantage over A*, which can already find the optimum solution. The small proportion of DQN outperformed cases in which DQN router outperforms in No.8 experiments (Type I dominant) originate from the randomness of problems generator: even with less nets, there will occasionally be some Type II problems generated.

However, when solving for a Type II problem, some edges will be fully utilized at a certain stage of the routing process when a portion of the nets have been routed. Therefore, the partial-edge-depletion capacity condition similar to what happens in Fig.~\ref{capacityExplain}e and Fig.~\ref{capacityExplain}f will take place during the routing process. In a partial-edge-depletion scenario, A*  will still solve each two-pin problems sequentially, without taking into consideration the change of the capacity conditions. In solving such problems,  DQN-based router is guided by the reward and solving the problem in an iterative training process. Furthermore, the capacity information is also encoded into the state representation of the DQN model. In this way, DQN router is able to solve a routing problem conjointly taking into account all nets and pins that need to be routed, which is described as the conjoint optimization mechanism.      

A comparison between the log-plot of reward curves when solving a Type I problem (Fig.~\ref{dqnfavor}d) and a Type II problem (Fig.~\ref{dqnfavor}h) provides  indirect evidence on how the conjoint optimization mechanism of DQN  works. In (Fig.~\ref{dqnfavor}d), which consists of ten randomly selected log-plot reward curves when solving Type I problems, the reward increases significantly in the first few hundred episodes. After that, it increases slowly and reaches a plateau gradually. For the ten log-plot reward curves for the Type II problems, although these curves have also ramped up significantly in the first few hundred episodes, they keep oscillating throughout the training process. This corresponds to the process of model weights adjustment in order to  ``reroute'' existing solutions, which  can yield better solutions by circumventing congested areas. 

\begin{figure}[t]
\centering
\includegraphics[scale=0.55]{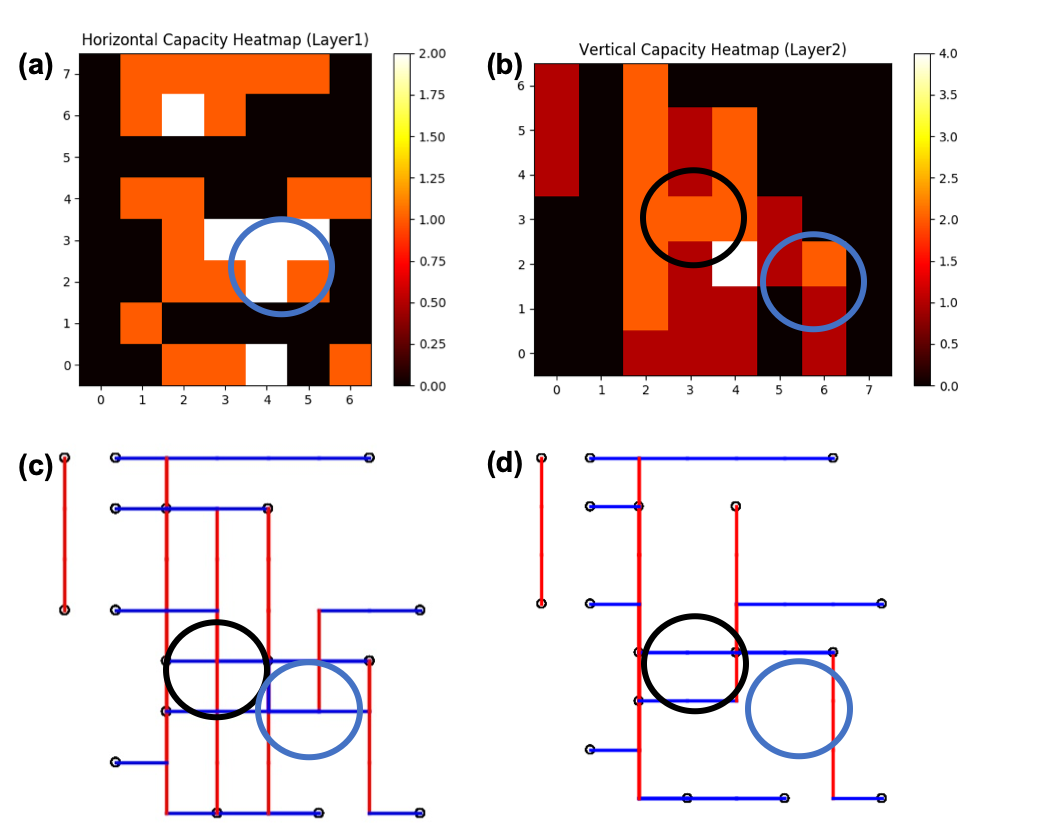}
\caption{Case study showing the conjoint optimization  of DQN router: heat map showing capacity utilization on (a) horizontal direction and (b) vertical direction based on A* routing, (c) A* routing solution, (d) DQN routing solution (red: layer 2, blue: layer 1)}
\label{casestudyDRLA} 
\end{figure}

A more direct case study showing the conjoint optimization  of the DQN router is based on the solution analysis and comparison of A* and DRL  on a toy problem, consisting of 10 nets (each one with a maximum of 2 pins) on an $8\times 8\times 2$ space. In solving this toy problem, DRL router with A*-based burn-in memory is used, $\gamma $ is set to be 0.9 and $\epsilon$ is set to be 0.05. Fig.~\ref{casestudyDRLA} shows some of the results for the case study. Fig.~\ref{casestudyDRLA}a and Fig.~\ref{casestudyDRLA}b are the heat maps showing edge utilization in the horizontal and vertical directions based on A* routing solution. The via capacity heat map is not displayed here since in the setting of this work, via capacity is set to be high enough that no OF will happen. The exact solution of A* routing is shown in  Fig.~\ref{casestudyDRLA}c, which has WL of 94 and OF of 1. The solution for exactly the same problem based on DQN router is shown in Fig.~\ref{casestudyDRLA}d, which has WL of 92 and OF of 0. By comparing the solutions of two routers, it is obvious that in the DQN router  there are some reduced connections in both the vertical direction (in the purple and black circle) and the horizontal direction (in the purple circle). From the heat maps of A* routing, it can be seen that the reduced-connection areas correspond to edges that are more actively utilized, or in other words more congested. Such an optimization procedure is not hard-coded, but instead realized by the conjoint optimization  of the DQN router. 

The major reason why DQN conjointly optimizes the problem is that the DQN model solves a whole routing problem with a single model, thus taking into account all the subtasks (nets and pins to be connected in this work) simultaneously. For the A* router, each subtask is solved individually and in this way the results will be strongly influenced by the order in which the subtasks are solved. In effect, the pre-routed nets will strongly influence the quality of routing  for pins and nets routed later. Although there can be  heuristic based algorithms to order the subtasks to combat this issue, such as routing nets with greater spanning areas first, such heuristics may not generalize well for all problems. For example, a spanning area based net ordering heuristics will be highly ineffective when solving a routing problem in which the spanning areas of most nets are similar.

\subsection{DQN Model Fine Tuning}

% XXX Add descriptions in the captions on what is the design variable for each of the plots. 
%[yyyexplain] : (above suggestion done)

% \begin{figure*}[thpb]
% \centering
% \includegraphics[trim = Oin Oin Oin Oin, clip,
% width=\textwidth]{FigTable/Fig_BatchExplore.png}
% \caption{REWARD PLOT BASED ON DRL MODELS WITH DIFFERENT BATCH SIZE}
% \label{bacthexplore} 
% \end{figure*}

% \begin{figure*}[thpb]
% \centering
% \includegraphics[trim = Oin Oin Oin Oin, clip,
% width=\textwidth]{FigTable/Fig_LearningRateExplore.png}
% \caption{REWARD PLOT BASED ON DRL MODELS WITH DIFFERENT LEARNING RATE}
% \label{learningrateexplore} 
% \end{figure*}

% \begin{figure*}[thpb]
% \centering
% \includegraphics[trim = Oin Oin Oin Oin, clip,
% width=\textwidth]{FigTable/Fig_NetworkExplore.png}
% \caption{REWARD PLOT BASED ON DRL MODELS WITH DIFFERENT NETWORK STRUCTURES}
% \label{networkexplore} 
% \end{figure*}

% \begin{figure*}[thpb]
% \centering
% \includegraphics[trim = Oin Oin Oin Oin, clip,
% width=\textwidth]{FigTable/Fig_EpsilonExplore.png}
% \caption{REWARD PLOT BASED ON DRL MODELS WITH DIFFERENT VALUES OF $\epsilon$}
% \label{epsilonexplore} 
% \end{figure*}

% XXX move the captions in each sub-figure to the captions of the whole figure. Add (a), (b), (c) to specify

In order to obtain DQN models with enhanced performance, fine tuning of the model is conducted in two steps: \textbf{general parameter tuning} and \textbf{performance improvement tuning}. General parameter tuning aims at stabilizing the training process and guarantee that there will be feasible solutions generated at the end of  training, which includes: batch size, learning rate, network structures and value of $\epsilon$. After a general parameter tuning, a model performance tuning is conducted on further increasing the model's ability to yield more desirable routing solutions, which also consists of a decay $\epsilon$ strategy, $\gamma$. Finally, the choice of memory burn-in is studied and analyzed.

\subsubsection{General Parameter Tuning}
The performance of different models is  evaluated in terms of reward evolution during DQN model training on $4\times 4\times 2$ toy problems with only 3 nets (each one with a maximum of 2 pins) and the training episodes is 2000. The simplified setting for general parameter tuning is based on our experimental results indicating the rough region of parameters that have a reasonable performance on most problems. The plotted rewards are not cumulative but the rewards summation for an episode, i.e. rewards summation for all two-pin routing sub-task in a problem. The rewards summation in an episode thus is obtained from the summation of rewards over all two-pin routing problems.

\begin{figure}[t]
\includegraphics[scale=0.26]{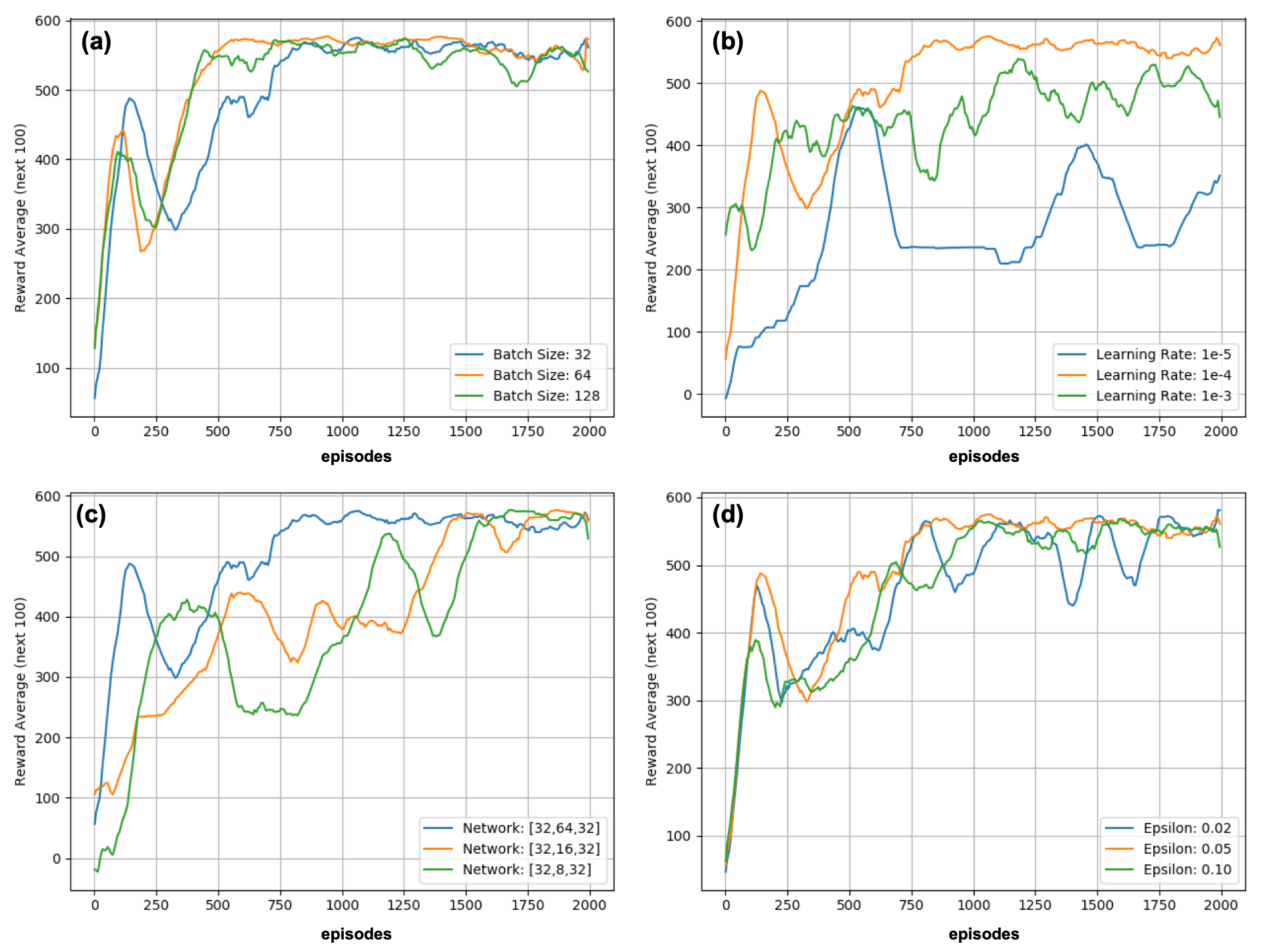}
\caption{Reward plots based on different combinations of model parameters. (a) batch size, (b) learning rate, (c) network structures, (d) epsilon.}
\label{parametricstudy} 
\end{figure}

\textbf{Batch size:} Batch sizes of 32, 64 and 128 are studied and the reward plots of models across three different batch sizes are shown in Fig.~\ref{parametricstudy}. The results indicate that increased batch size does not cause the reward to increase faster and could lead to oscillations in the reward when the batch size is 128. 

\textbf{Learning rate:} Experimented learning rates include 1e-3, 1e-4 and 1e-5. These produce a marked difference in the reward evolution during training as shown in Fig.~\ref{parametricstudy}. When the learning rate is  small, the reward increase is slow and unstable. After 2000 episodes, the model fails to produce a reasonable solution. A more feasible learning rate typically already produces  feasible solutions after 2000 episodes. On the other hand, when the learning rate is too large, strong oscillations exist throughout the whole training. A learning rate of 1e-4 results in the best trade-off.

% wwww: change all loops and episodes back to episodes. This includes the figures. For instance Figure 6.
% [wwwresponse]:(done)

\textbf{Network structures:} Various parameters including the number of layers, the number of nodes each layer and activation functions are studied. In particular, three fully-connected networks with different number of nodes in their second layer are investigated. Fig.~\ref{parametricstudy} shows the reward plot of DRL models with three different networks. As shown, decreasing the node numbers of the second layer  negatively influences the performance of DQN models by increasing the time it takes to obtain a higher reward or making the learning process unstable. 

\textbf{$\epsilon$ value:} As shown in Eqn.~\ref{epsilongreedyeqn}, the value of $\epsilon$ controls the exploration-exploitation rate with higher $\epsilon$ values corresponding to more randomized actions. Fig.~\ref{parametricstudy} shows the reward plots based on different $\epsilon$ values. When $\epsilon$ is 0.05, the reward evolution is more stable compared to cases in which $\epsilon$ is smaller or greater than 0.05. Besides, a commonly adopted decay $\epsilon$ strategy is experimented in the performance improvement tuning. 

Based on the above experimentation, our optimized parameters  are as follows. Batch size: 32, learning rate: 1e-4, nodes in  layers: [32, 64, 32], $\epsilon$: 0.05. While the best parameter set may vary with problem complexity such as the grid size, the number of nets and the number of pin pairs on a grid, our own experience indicates that the complexity of the network structure needs to be increased to accommodate the  increase in  problem complexity. However, a quantitative approach that reveals the best network structure as a function of the given model size without overfitting is the subject of our future studies.

\subsubsection{Performance Improvement Tuning}

\begin{figure*}[thpb]
\centering
\includegraphics[width=\textwidth]{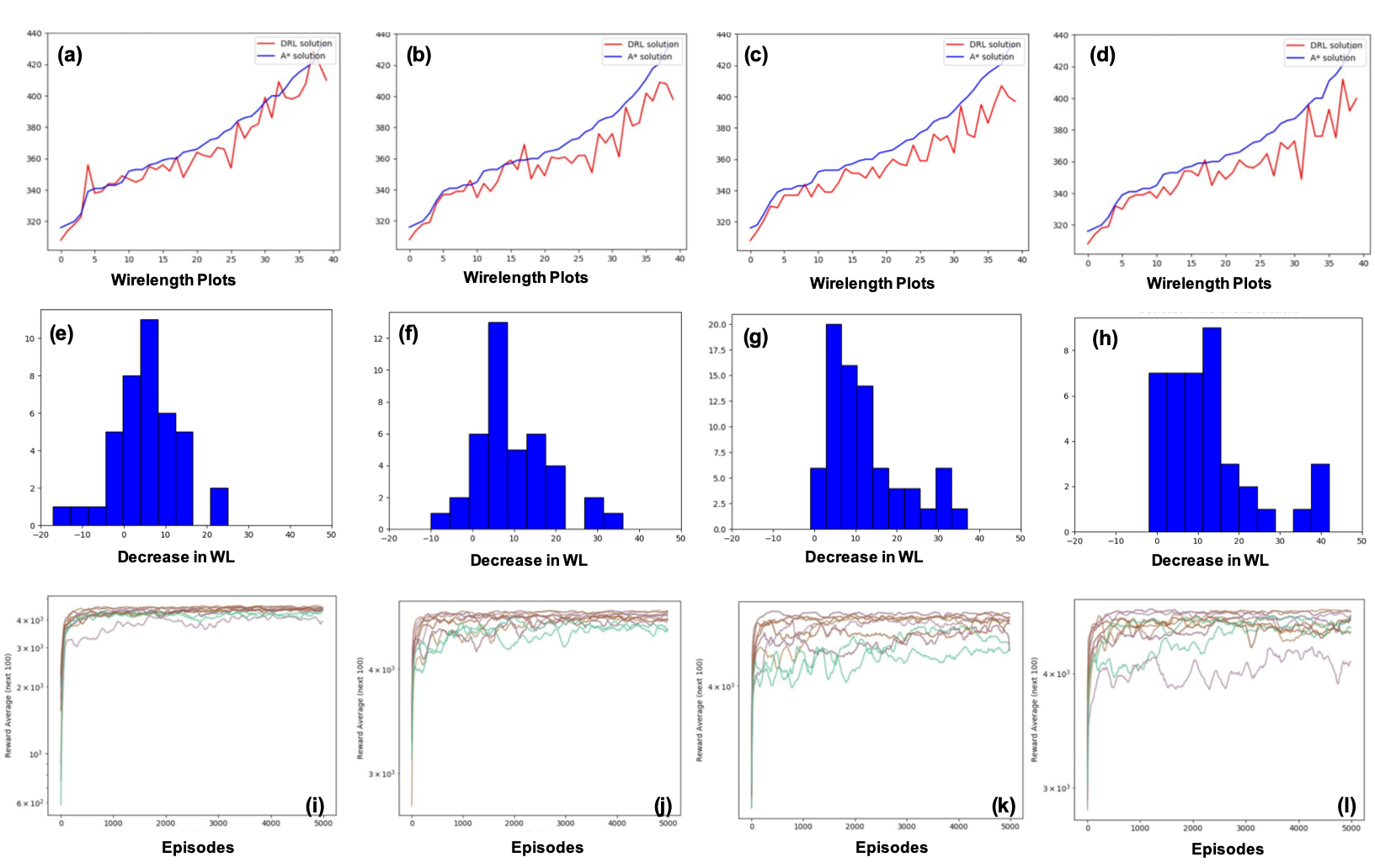}
\caption{Influence of discount factor $\gamma$ on performance of DQN router and reward curves. WL of DQN routing solutions (red) with $\gamma$ value of (a) 1, (b) 0.95, (c) 0.9, (d) 0.8 and their comparison to A* routing solutions (blue); Decrease in WL of DQN-based routing solutions compared to A* routing solutions with  $\gamma$ value of (e) 1, (f) 0.95, (g) 0.9, (h) 0.8; log-plot of reward curves for 10 randomly selected problems solved by DQN router with $\gamma$ value of (i) 1, (j) 0.95, (k) 0.9, (l) 0.8. [To do: e,f,g,h: same x scale]}
\label{gammatuning} 
\end{figure*}

Based on the previous general parameter tuning, a tuning aimed at improving the model performance for solving routing problems is conducted, focusing on $\gamma$ and $\epsilon$, which will significantly influence the model performance. Firstly a decay $\epsilon$ strategy, which linearly decreases $\epsilon$ from 0.05 to 0.01 starting from 2000 episodes, is experimented in experiment No.7 shown in Table.~\ref{exp_num}. Compared with experiment No. 5 which does not include decay $\epsilon$ strategy, there is a minor decrease in the performance. Therefore, the decay $\epsilon$ strategy is not adopted in the optimized models.

Then, models based on different values of $\gamma$ (1, 0.95, 0.9, 0.8), which is the discount factor determining the trade-off between short term and long term reward, are experimented to solve 40 problems with parameters specified in No.2, No.3, No.4 and No.5. In this research, short term reward  corresponds to the negative reward obtained from a random walk, while long term reward corresponds to the positive reward obtained when the target pin is reached. Fig.~\ref{gammatuning} shows the statistical results for routing solutions of DQN-based router with different values of $\gamma$. Fig.~\ref{gammatuning}a, Fig.~\ref{gammatuning}b, Fig.~\ref{gammatuning}c and Fig.~\ref{gammatuning}d show the WL for the 40 experimented problems of DQN routing solutions (red) with $\gamma$ values of 1, 0.95, 0.9, 0.8 and their corresponding A* routing solutions (blue). In most cases, DQN router gives better solutions with less WL in and the percentage of problems in which DQN router yields better solutions are 80\% ($\gamma$=1), 90\%($\gamma$=0.95), 97.5\%($\gamma$=0.9) and 95\%($\gamma$=0.8) respectively. Thus, the optimum $\gamma$ based on the statistical results are chosen to be 0.9. Histograms in  Fig.~\ref{gammatuning}e, Fig.~\ref{gammatuning}f, Fig.~\ref{gammatuning}g and Fig.~\ref{gammatuning}h shows the decrease in WL of DQN-based routing solutions on the 40 experimented problems compared to A*-based routing with $\gamma$ values of 1, 0.95, 0.9, 0.8. The one (Fig.~\ref{gammatuning}g) that corresponds to best $\gamma$ shows that on all the 40 problems experimented, WL gets non-negative decrease. It is worth mentioning that in these 40 problems, a small proportion of problems had a non-zero OF based on A* routing. However, in all cases, the OF of DQN routing solutions is zero. The above analysis indicates the overwhelmingly better performance of DQN router compared with A* router on Type II problems. 

The log-plot  of reward curves for 10 randomly selected problems solved by DQN router with $\gamma$ value of 1, 0.95, 0.9, 0.8 are shown in  Fig.~\ref{gammatuning}i, Fig.~\ref{gammatuning}j, Fig.~\ref{gammatuning}k and Fig.~\ref{gammatuning}l. The first observation is that oscillations of reward curves increased considerably as $\gamma$ increases, indicating greater adjustment of DQN-network weights to obtain conjointly optimized solutions. The reason for such a change is that when $\gamma$ is too large and even close to one, the DQN model will no longer be sensitive to a change of WL represented by short term negative reward. Only when $\gamma$ becomes reasonably small and the short term negative reward starts having a greater influence over the expected total reward will the conjoint optimization  start to work. Another important observation is that when $\gamma$ becomes too small, especially when it is 0.8 (Fig.~\ref{gammatuning}l), some reward curves plateau at a lower level compared to where they would have reached when $\gamma$ is 1 (Fig.~\ref{gammatuning}i). One plausible explanation for the degeneration of solutions as $\gamma$ gets too small is that positive reward obtained from reaching a target pin diminishes when multiplied with multiple $\gamma$.

\subsubsection{Burn-In Memory Choice}
\begin{figure}[t]
\centering
\includegraphics[scale=0.6]{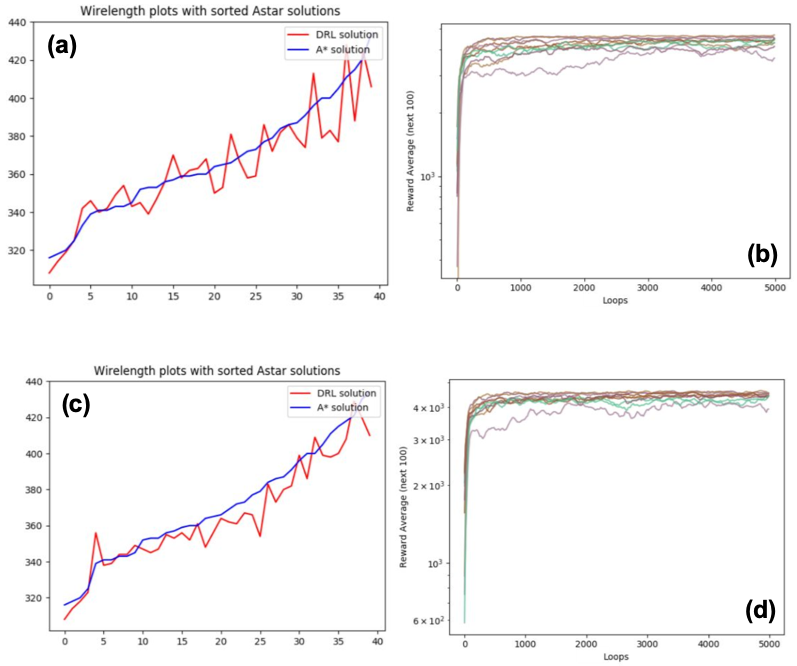}
\caption{Burn-in memory study showing statistical results of DQN router on Type II problems: WL of DQN-based routing solutions (red) with (a) random burn-in memory and (c) A* based burn-in memory and their comparison to A*-based routing solutions (blue); log-plot of reward curves for 10 randomly selected problems solved by DQN-based routing with (b) random burn-in memory and (d) A* based burn-in memory.}
\label{burninstudy} 
\end{figure}

The burn-in memory for the DQN model is yet another key component and the choice of burn-in memory is regarded as a tunable parameter. Here, a comparison of a random vs. A* based burn-in memory, which use A* routing solutions for the same problem, is conducted to determine the appropriate choice of burn-in memory. Experimental results corresponding to No. 5 and No. 6 in Table.~\ref{exp_num} demonstrate the performance of DQN router on Type II problems with A*-based and random memory burn-in respectively. Fig.~\ref{burninstudy}a and Fig.~\ref{burninstudy}c show the WL of DQN router with random and A* router memory burn-in respectively (red), with the corresponding WL of A* routing solutions (blue) also plotted for comparison. For random memory burn-in case, DQN routing solutions show no signs of advantage and only in 57.5\% cases have smaller WL compared with A* routing solutions. With the application of A* router memory burn-in, DQN routing solutions are mostly better than corresponding A* routing solutions. To be specific, in 80\% of the problems experimented, DQN router can outperform A* router. In terms of OF, DQN routing solutions all have zero OF while A* router can occasionally have positive OF. For the log-plot of reward curves of 10 randomly selected problems solved by DQN router with random (Fig.~\ref{burninstudy}b) and A* router (Fig.~\ref{burninstudy}d) based memory burn-in, the difference is minuscule. Thus, A* router memory burn-in is selected owing to its positive effect on performance improvement, yet it will not considerably increase the learning speed of models based on the experimental results of this research.

\subsection{DRL Solutions on Different Problems} 

\begin{figure}[t]
\begin{center}
\includegraphics[scale=0.36]{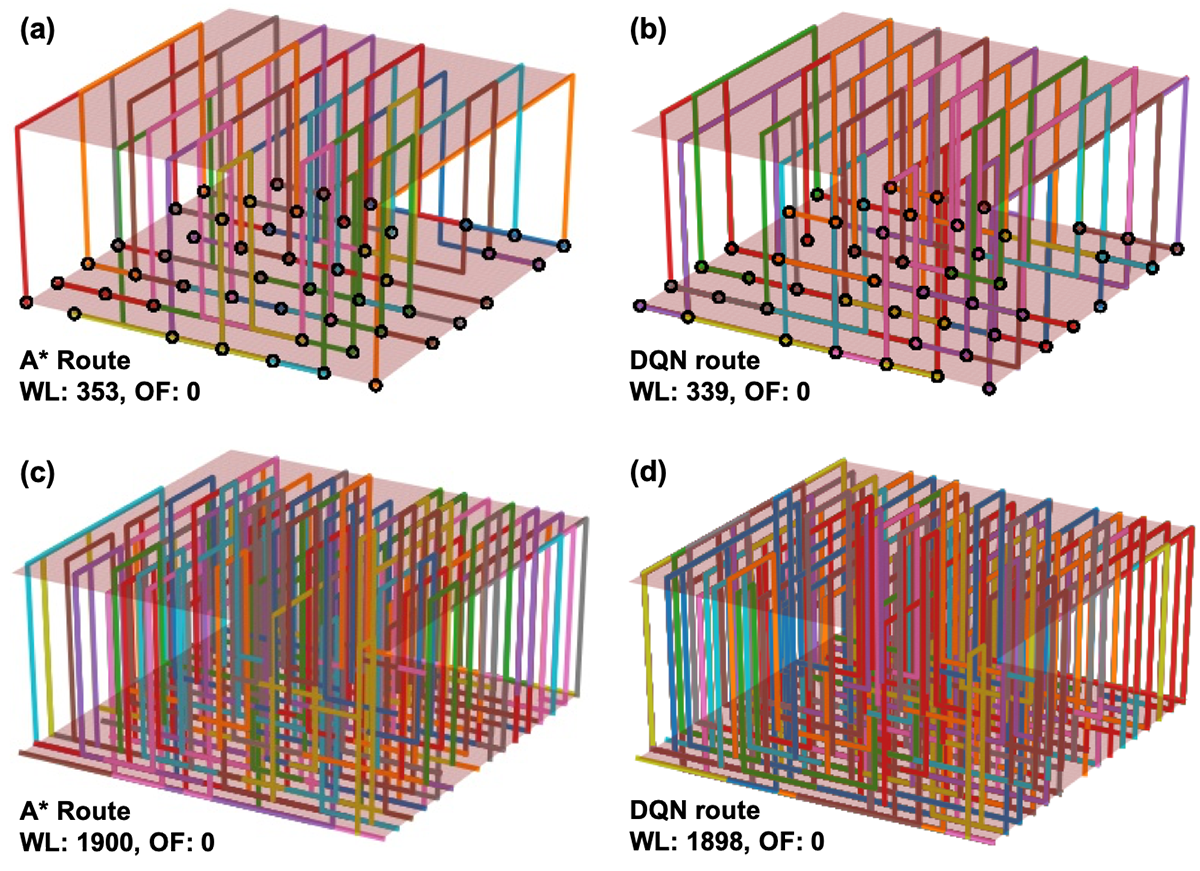}
\end{center}
\caption{Visualized results showing routing solutions for $8\times 8\times 2$ problem based on (a) A* router and (b) DQN router; $16\times 16\times 2$ problem based on (c) A* router and (d) DQN router.}
\label{8and16solutions} 
\end{figure}

Based on the optimized network structure and model parameters, the visualized solutions of DQN routing solutions on both $8\times 8\times 2$ and $16\times 16\times 2$ problems that are both Type II are presented in Fig.~\ref{8and16solutions} with their respective WL and OF. The results show that the tuned DQN model not only yields better solutions on $8\times 8\times 2$ scale problems but also performs better when directly applied to solving $16\times 16\times 2$ scale problems, without change of model parameters or structures.

%%%%%%%%%%%%%%%%%%%%%%%%%%%%%%%%%%%%%%%%%%%%%%%%%%%%%%%%%%%%%%%%%%%%%%
\section{CONCLUSIONS}

This work is the first attempt to formulate the historically challenging global routing as a deep reinforcement learning problem. By following the established conventions of solving problems sequentially and pin decomposition in global routing,  the proposed DQN router produces superior solutions over an A* router in most cases for Type II (partial-edge-depletion) problems, even though A* uses an admissible heuristic. The conjoint optimization that leads to the superior performance of DQN router is analyzed and validated experimentally. Parameters including batch size, learning rate, network structures, $\epsilon$, $\gamma$ and memory burn-in are fine-tuned to enhance the performance of the DQN model. A global routing problem sets generator is developed with the ability to generate global routing problem sets fast with user-defined size and constraints. These problem sets and code for the generator  will be used for training our DRL approach and will be made publicly available to the community.

\section{Future Work}

There are two primary ways the current work can be extended to make it applicable to bigger-sized problems.  Firstly, as pointed out in previous research, the conventional  deep Q-network adopted in this work tends to overestimate the value of the Q-function. Better models such as double deep Q-learning (double DQN) \cite{van2016deep} are promising alternatives for enhancing the performance. Secondly, only a single agent is used in our current approach, which does not naturally extend from global routing problems where each net often has more than two pins to be interconnected. Models with multiple agents \cite{multiagentDRL_review,hu1998multiagent,bu2008comprehensive,tan1993multi} are likely to perform better at a set of simultaneous multi-pin problems, hence alleviating the restriction for solving the problem sequentially and therefore yielding better solutions.  

%[yyyexplain]:(the most impressive advantage of multiagent approach is to free the algorithm from sequential approach, thus I made this revision by emphasizing sequentially)（imposed by  net ordering and multi-pin decomposition） 

Finally, a major obstacle in applying advanced DRL or other machine learning based algorithms to solve global routing problems is the absence of publicly available problem set repositories that include abundant global routing problems with varying sizes and constraints. In follow-up work, we intend to train our model based on a large number of automatically generated problem sets that will not only help enable the scaling of the proposed system to larger grids with more complex net configurations but also increase the model's generalization ability to solve unseen problems. Prioritized experience replay approach \cite{schaul2015prioritized} will be utilized to train the Q-network with important transitions more frequently.

\bibliographystyle{asmems4}

%%%%%%%%%%%%%%%%%%%%%%%%%%%%%%%%%%%%%%%%%%%%%%%%%%%%%%%%%%%%%%%%%%%%%%
\begin{acknowledgment}
This work is partially funded by the DARPA IDEA program (HR0011-18-3-0010). The authors would like to thank Rohan Jain, Jagmohan Singh, Elias Fallon and David White from Cadence Design Systems for the constructive feedback and the establishment of the simulated routing environment. 
\end{acknowledgment}

\newpage
\bibliography{asme2e}

\newpage
\appendix       %%% starting appendix
\section{Appendix A: Heat maps showing edge utilization of different problem sets based on A* solution}

\begin{figure*}[thpb]
\centering
\includegraphics[width=\textwidth]{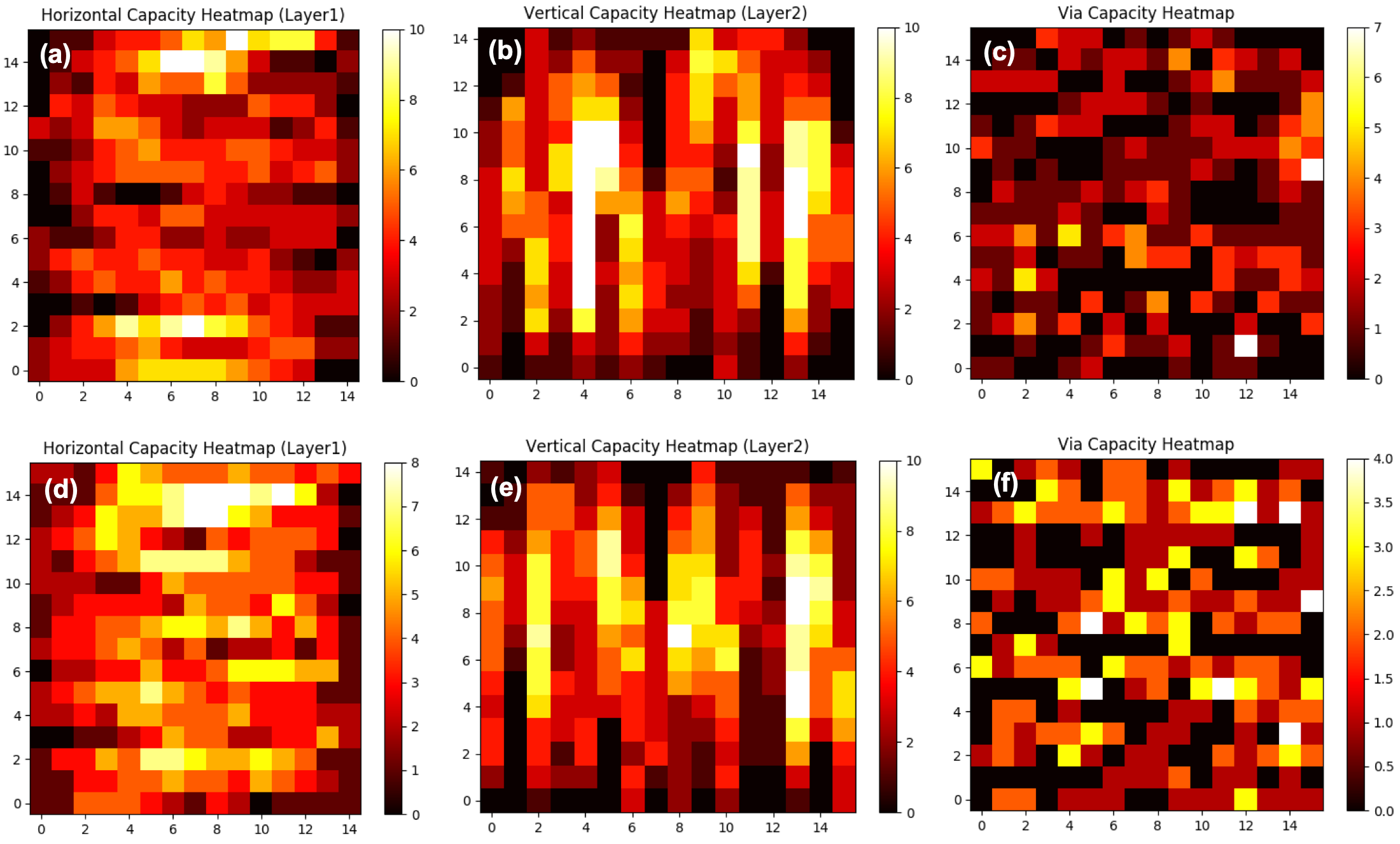}
\caption{Heat maps and edge utilization histogram of two global routing problems generated with global routing problem sets generator.  $16\times 16\times 2$ benchmark with normal capacity setting: (a) via capacity heat map, (b) vertical capacity heat map, (c) horizontal capacity heat map;  $16\times 16\times 2$ benchmark with reduced capacity setting: (d) via capacity heat map, (e) vertical capacity heat map, (f) horizontal capacity heat map (number of nets: 160, max number of pins each nets: 2, normal capacity: 3)}
\label{heat} 
\end{figure*}

\begin{figure*}[thpb]
\centering
\includegraphics[width=\textwidth]{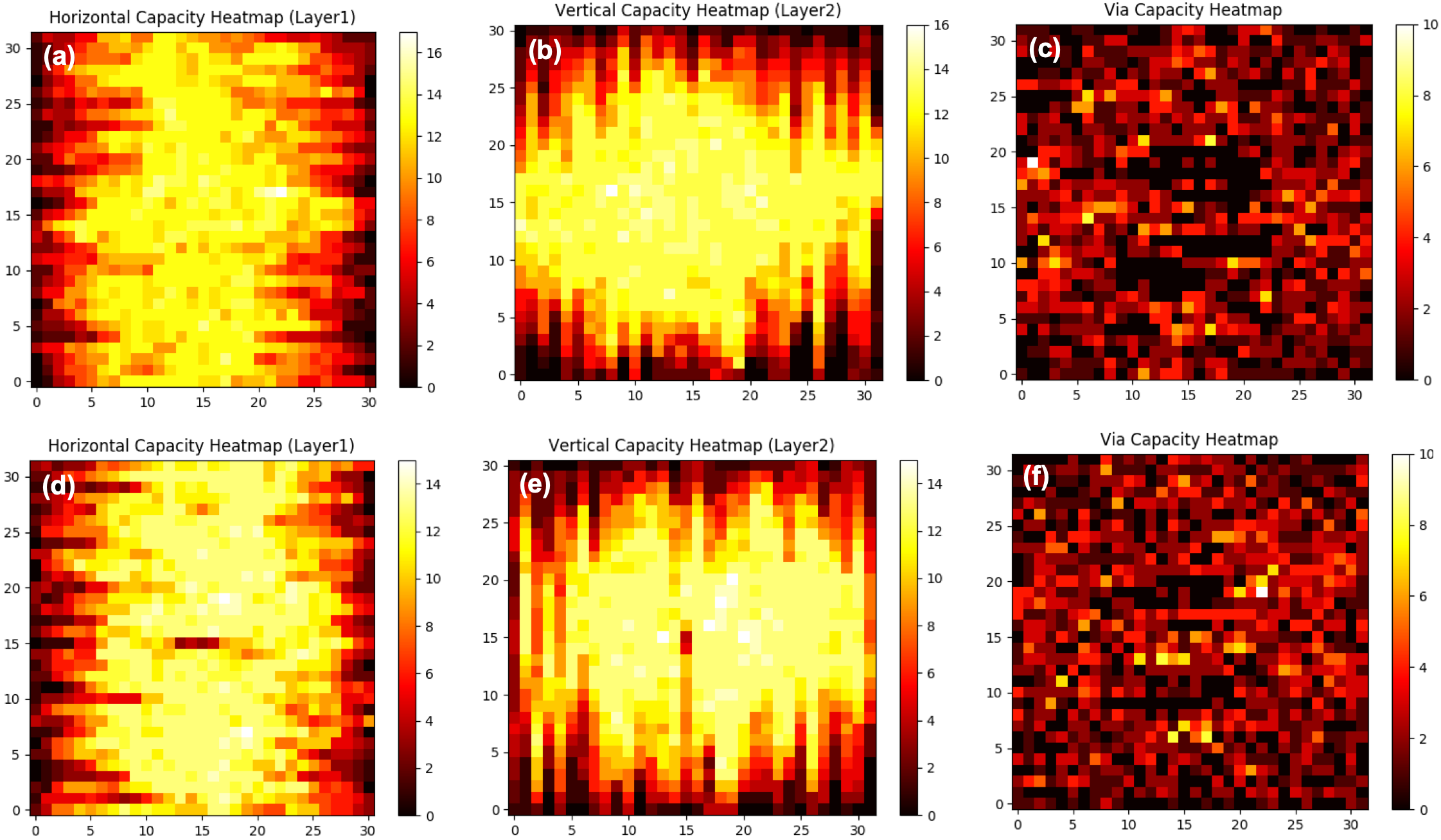}
\caption{Heat maps and edge utilization histogram of two global routing problems generated with global routing problem sets generator.  $32\times 32\times 2$ benchmark with normal capacity setting: (a) via capacity heat map, (b) vertical capacity heat map, (c) horizontal capacity heat map;  $32\times 32\times 2$ benchmark with reduced capacity setting: (d) via capacity heat map, (e) vertical capacity heat map, (f) horizontal capacity heat map (number of nets: 160, max number of pins each nets: 2, normal capacity: 3)}
\label{heat} 
\end{figure*}

% \begin{figure*}[thpb]
% \centering
% \includegraphics[width=\textwidth]{FigTable/Fig_64bench.png}
% \caption{Heat maps and edge utilization histogram of two global routing problems generated with global routing problem sets generator.  $64\times 64\times 2$ benchmark with normal capacity setting: (a) via capacity heat map, (b) vertical capacity heat map, (c) horizontal capacity heat map;  $64\times 64\times 2$ benchmark with reduced capacity setting: (d) via capacity heat map, (e) vertical capacity heat map, (f) horizontal capacity heat map; $64\times 64\times 2$ benchmark with normal capacity but 100 nets: (g) via capacity heat map, (h) vertical capacity heat map, (i) horizontal capacity heat map; (number of nets: 30, max number of pins each nets: 12, normal capacity: 3)}
% \label{heat} 
% \end{figure*}

\section{Appendix B: DQN Global Routing Algorithm}
\hrule
\vspace{3pt}
\noindent 1. Burn in replay memory D=$e_1,e_2,...,e_t$ to capacity N\\
$e_t = (s_t,a_t,r_t,s_{t+1}, IsTerminal_{t+1})$  \\
2. Initialize action-value function Q with random weights \\
3. Set batch size: n\\
4. Decompose multi-pin problems with Minimum Spanning Tree\\
5. Network training:\\
\textbf{for} episode = 1, maximum episodes\\
\indent Initialize sequence $s_1$ starting at one pin\\
\indent With probability $\epsilon$ select a random action $a_t$\\
\indent otherwise select $a_t = max_a Q^*(\phi(s_t,a;\theta))$ \\
\indent Execute action $a_t$ in environment and observe reward $r_t$\\
\indent Update the \textbf{environment} (capacity)\\
\indent \textbf{for} t = 1, maximum steps\\
\indent \indent Set $s_{t+1} = s_t,a_t$\\ 
\indent \indent Store transition $(s_t,a_t,r_t,s_{t+1}, IsTerminal_{t+1})$ in D\\
\indent \indent Sample random minibatch of transitions from D\\
\indent \indent  Set $y_j=
\begin{cases}
r_j & \text{for terminal} s_{j+1}\\
r_j+\gamma max_{a'}Q(s_{j+1},a';\theta) & \text{for non terminal }  s_{j+1}
\end{cases}$\\
\indent \indent Perform a gradient descent step on $(y_i-Q(s_j,a_j;\theta))^2$\\
\indent\textbf{end for}\\
\textbf{end for}

\vspace{3pt}

\bibliographystyle{asmems4}

\hrule

\end{document}